%% file: paper.tex
\documentclass[format=acmsmall,screen=true, prologue, table]{acmart}
\pdfoutput=1

\usepackage{booktabs}

\acmJournal{TKDD}
\acmVolume{1}
\acmNumber{1}
\acmArticle{1}
\acmYear{2023}
\acmMonth{6}

\input{preamble}

\usepackage{subfigure}
\usepackage{graphbox}
\usepackage{svg}
\usepackage{nicefrac}

\usepackage{bold-extra}

\usepackage{array}
\newcolumntype{P}[1]{>{\centering\arraybackslash}p{#1}}
\newcolumntype{M}[1]{>{\centering\arraybackslash}m{#1}}
\newcolumntype{R}[1]{>{\arraybackslash}m{#1}}

\definecolor{orange}{rgb}{1,0.5,0}
\definecolor{graynode}{RGB}{20,20,20}
\definecolor{crimsonred}{RGB}{220,20,60}
\definecolor{darkgraynode}{gray}{0.5}
\definecolor{lightgraynode}{gray}{0.8}

\usepackage{rotate}
\usepackage{capt-of}
\usepackage{tabulary}
\usepackage{setspace}
\usepackage{mathtools}
\usepackage{pifont} 
\newcommand{\cmark}{\ding{51}}
\newcommand{\xmark}{\ding{55}}

\definecolor{gray}{RGB}{20,20,20}
\definecolor{gray}{RGB}{0.7,0.7,0.7}
\definecolor{greencm}{RGB}{0,153,0}

\newcommand{\cm}{ {\color{greencm}\normalsize\cmark}}
\newcommand{\cmgray}{ {\color{gray}\normalsize\cmark}}
\newcommand{\xm}{ {\color{red}\normalsize\xmark}}

\definecolor{plotblue}{RGB}	{30,144,255}
\definecolor{plotgreen}{RGB}	{50,205,50}
\definecolor{plotred}{RGB}	{220,20,60}

\definecolor{myyellow}{RGB}{255,255,204}
\definecolor{myred}{RGB}{255,204,204}
\definecolor{myblue}{RGB}{0,200,255}
\definecolor{mygreen}{RGB}{80,220,80}

\newcommand*\hrulefillvar[1][0.4pt]{\leavevmode\leaders\hrule height#1\hfill\kern0pt}

\usepackage{multirow}

\definecolor{thedarkblue}{RGB}{0,0,120} 
\definecolor{mydarkblue}{rgb}{0,0.08,0.45} 

\usepackage{hyperref}
\hypersetup{%
    colorlinks=true,
    linkcolor=mydarkblue,
    citecolor=mydarkblue,
    filecolor=mydarkblue,
    urlcolor=mydarkblue}

\DeclareMathAlphabet{\mathbcal}{OMS}{cmsy}{b}{n}
\usepackage{amsmath}
\usepackage{mathrsfs}
\usepackage{comment}

\usepackage{bm}
\usepackage{bbm}
\usepackage{enumitem}
\AtBeginDocument{
  \providecommand\BibTeX{{
    \normalfont B\kern-0.5em{\scshape i\kern-0.25em b}\kern-0.8em\TeX}}}

\begin{document}
\title{Fairness-Aware Graph Neural Networks: A Survey}

\settopmatter{authorsperrow=3}

\author{April Chen}
\affiliation{%
  \institution{Harvard University}
}
\email{aprilchen@college.harvard.edu}

\author{Ryan A. Rossi}
\affiliation{
  \institution{Adobe Research}
}
\email{rrossi@adobe.com}

\author{Namyong Park}
\affiliation{
  \institution{Meta AI}
}
\email{namyongp@meta.com}

\author{Puja Trivedi}
\affiliation{
  \institution{University of Michigan}
}
\email{pujat@umich.edu}

\author{Yu Wang}
\affiliation{
  \institution{Vanderbilt University}
}
\email{yu.wang.1@vanderbilt.edu}

\author{Tong Yu}
\affiliation{
  \institution{Adobe Research}
}
\email{tyu@adobe.com}

\author{Sungchul Kim}
\affiliation{
  \institution{Adobe Research}
}
\email{sukim@adobe.com}

\author{Franck Dernoncourt}
\affiliation{
  \institution{Adobe Research}
}
\email{dernonco@adobe.com}

\author{Nesreen K. Ahmed}
\affiliation{
  \institution{Intel Labs}
}
\email{nesreen.k.ahmed@intel.com}

\renewcommand{\shortauthors}{Chen et al.}

\begin{abstract}
Graph Neural Networks (GNNs) have become increasingly important due to their representational power and state-of-the-art predictive performance on many fundamental learning tasks. Despite this success, GNNs suffer from fairness issues that arise as a result of the underlying graph data and the fundamental aggregation mechanism that lies at the heart of the large class of GNN models. In this article, we examine and categorize fairness techniques for improving the fairness of GNNs. Previous work on fair GNN models and techniques are discussed in terms of whether they focus on improving fairness during a preprocessing step, during training, or in a post-processing phase. Furthermore, we discuss how such techniques can be used together whenever appropriate, and highlight the advantages and intuition as well. We also introduce an intuitive taxonomy for fairness evaluation metrics including graph-level fairness, neighborhood-level fairness, embedding-level fairness, and prediction-level fairness metrics. In addition, graph datasets that are useful for benchmarking the fairness of GNN models are summarized succinctly. Finally, we highlight key open problems and challenges that remain to be addressed.
\end{abstract}



\maketitle

\section{Introduction}
Graph Neural Networks (GNNs) have become increasingly important in recent years and successfully used in many areas since many real-world data are represented as graphs, such as social networks, financial data, and molecular data. 
Despite the initial success of GNNs, learning fair and unbiased GNN models remains a fundamentally important and challenging problem.
Due to the inherent nature of GNNs and how they leverage the graph structure to perform aggregation over neighborhoods, ensuring fairness and controlling the accuracy and fairness trade-off is significantly more difficult compared to non-graph models that use i.i.d. data.

Since neighborhoods lie at the heart of all GNN-based methods~\cite{wu2020comprehensive}, the fairness of the trained GNN models and the resulting embeddings learned are fundamentally tied to the neighborhoods used to iteratively train these models~\cite{zhang2023interaction,chen2022FairNeigh,hussain2022adversarial}.
Furthermore, it is often very difficult to design GNNs that mitigate such unfairness and bias issues that arise due to the nature of the graph structure, input features, and most importantly, the fundamental GNN assumptions and design that make this a far more challenging and complex problem compared to traditional bias and unfairness mitigation techniques for i.i.d. data~\cite{song2022guide,xu2023gfairhint,salganik2022analyzing}.
In fact, GNNs are largely designed to leverage such bias and unfairness in the data to achieve superior accuracy at the expense of fairness~\cite{kose2022fairnorm,singer2022eqgnn,kose2022fairness,kose2022fair,liu2023fair,dong2023reliant,kose2022fast,wang2023fair}.
As an example, a GNN-based recommender may suggest fewer job opportunities to individuals of a specific gender or ethnic group.
This is due to the fact that most graph data is highly skewed towards one or more groups and often even shows a rough power-law relationship as observed in the literature across a variety of domains in the last decade~\cite{newman2003structure,watts1998collective}.
Therefore, fairness in such models are both practically and theoretically important to develop better GNN models that are significantly more fair while also accurate~\cite{liu2022trade} for downstream prediction tasks such as node classification~\cite{ma2021subgroup,loveland2022graph,agarwal2021towards,zhu2023fairness}, link prediction~\cite{pmlr-v119-buyl20a,FairAdj,patro2020fairrec,FairDrop,rahman2019fairwalk}, and link classification~\cite{chen2022FairNeigh}.

In this work, we discuss the fairness issues that arise in GNNs and survey the techniques to improve fairness in GNNs.
We highlight three fundamental facets that can lead to bias when training GNN models.
First, the underlying graph structure $G$ used for training is often biased, $\eg$, when considering an attribute of a node (representing an individual) such as political views, we often observe significant homophily among the neighbors of nodes. 
In fact, in such data, there are often very tightly-knit communities of individuals that all retweet or follow each other.
Second, the features given as input to GNNs can also be biased and unfair in a variety of ways. 
Such features when used independently may essentially have all the unfairness issues of traditional i.i.d. data.
Third, the underlying mechanism used for aggregation and training of GNNs is inherently biased, and this is a much more difficult issue to resolve compared to traditional fairness on i.i.d. data.
Overall, fairness issues in GNNs arise due to various factors such as biased training data including both the input features along with the graph structure, as well as the training and aggregation mechanisms that lie at the heart of GNNs. Addressing these issues requires careful consideration of the data, model, and evaluation metrics to ensure fair and unbiased predictions.

\subsection{Summary of main contributions}
The key contributions of this work are as follows:
\begin{enumerate}
\item A comprehensive survey of existing work on bias and unfairness mitigation techniques for GNNs. We also survey graph fairness metrics and summarize existing graph datasets used in the literature by the domain the graph originates (\eg, social network) along with task it can be used for and the dataset statistics and characteristics useful for various graph settings.
\item We introduce a few intuitive taxonomies for bias mitigation in GNNs and survey existing methods using these taxonomies. 
The taxonomy categorizes techniques based on whether the approach mitigates unfairness at the pre-processing stage, training stage, or at the post-processing stage by debiasing the learned embeddings directly. 
Methods are also categorized by the type of input graph data supported such as whether the graph is homogeneous, bipartite, heterogeneous, or temporal, as well as by the underlying graph learning task for which the method was designed.
\item We identify key open problems and challenges that are important for future work to address in this rapidly emerging but critically important field.
\end{enumerate}

\subsection{Scope of this article}
In this article, we focus on examining and categorizing various fairness techniques for graph neural networks.
We do not attempt to survey the abundance of work on fairness in graph mining~\cite{dong2022fairness,zhang2022fairness} and graph machine learning in general~\cite{choudhary2022survey}.
In contrast, we focus solely on fair GNN models as opposed to general graph fairness.
In some cases, techniques we survey may have been used in a different context.
Regardless of the context, we examine the general applicability and benefits of these techniques when used for improving fairness in GNN models.

\section{Problem Formulation}
Given a graph $G=(V,E)$ consisting of a set of nodes $V$ along with a set of edges $E \subseteq V \times V$ that encode dependencies between pairs of nodes in $V$. Furthermore, every node $v \in V$ typically has a $k$-dimensional feature vector $\vx_v$ associated with it. 
This can be represented compactly as a node feature matrix $\mX = [\, \vx_1 \; \vx_2 \, \cdots\, \vx_{|V|} \,]^{\top} \in \RR^{|V| \times k}$. 
We also have one or more sensitive attributes $\vs = \big[\,s_1\; s_2\, \cdots\, s_i\, \cdots\, s_{|V|}\,\big]$ where $s_i$ is the sensitive attribute value of node $i$.
The graph is encoded as a sparse adjacency matrix $\mA$ where $A_{vu}=1$ if $(v,u) \in E$ and $A_{vu}=0$ otherwise. 
GNN functions operate over the local neighborhoods of the nodes in the graph, that is, the neighborhood $N_v$ for node $v$ is defined as 
$N_v = \{u \in V \,|\, (v,u) \in E\}$.
Hence, $N_v$ is the set of nodes adjacent to $v$.
From $N_v$, we define the multiset of features from the neighborhood of $v$ as $\mX_{N_v} = \{\!\!\{ \vx_{u} \,|\, u \in N_v \}\!\!\}$.

A key challenge of ensuring fairness in this setting with respect to the sensitive attribute $\vs$ is that this sensitive information is often encoded in the graph's adjacency matrix $\mA$ and even the feature matrix $\mX$. Both $\mA$ and $\mX$ are fundamental to the training of GNNs. 
In terms of the graph structure $\mA$, this occurs when the sensitive attribute values of the neighborhood $N_v$ of a node $v$ are overwhelming the same.
This implies the presence of homophily in $G$ where nodes sharing the same sensitive attribute value are more likely to be connected.
Conversely, the feature matrix $\mX$ may also be highly correlated with the sensitive attribute $\vs$, especially when diffused over the graph structure $\mA$:
\begin{align}\label{eq:gnn-layer-basic}
   \mH = \textsc{GNN}(\mX, \mA)
\end{align}
where GNN can be any GNN layer.
More formally, let $\phi$ denote a local diffusion (propagation) function that operates over the neighborhood of a node.
Then for node $v \in V$, we have 
$\vh_v = \phi(\vx_v, \mX_{N_v})$.
The majority of GNNs can be categorized into convolutional (Eq.~\ref{eq:conv}), attentional (Eq.~\ref{eq:atten}), or message-passing (Eq.~\ref{eq:msg-passing})~\cite{bronstein2021geometric}. In particular, 
\begin{align}
    \vh_v &= \phi\Bigg( \vx_v \bigoplus_{u \in N_v} c_{uv} \;\psi(\vx_u) \Bigg) \qquad \qquad\;\;\;\; \text{(Convolutional)} \label{eq:conv} \\
    \vh_v &= \phi\Bigg( \vx_v \bigoplus_{u \in N_v} a(\vx_u, \vx_v) \;\psi(\vx_u) \Bigg) \qquad \qquad \text{(Attentional)} \label{eq:atten} \\
    \vh_v &= \phi\Bigg( \vx_v \bigoplus_{u \in N_v} \psi(\vx_u, \vx_v) \Bigg) \qquad \qquad\; \text{(Message-passing)} \label{eq:msg-passing}
\end{align}
where $\psi$ and $\phi$ are neural networks (\eg, ReLU) and $\bigoplus$ is any aggregator such as $\sum$, mean, max, among others~\cite{rossi2018deep}. 

The fair GNN learning problem is to learn a low-dimensional fair embedding matrix $\mH \in \RR^{n \times d}$ of the nodes such that $d \ll n$. Most importantly, the embeddings must encode different properties of the graph structure along with the input features. 
Typically, it is assumed that two nodes connected in the graph have similar embeddings.
Furthermore, the embeddings $\mH$ must be independent of the sensitive attributes $\vs$ such that no information is revealed about the sensitive attribute $\vs$ from the learned embeddings $\mH$.
This problem is often very challenging since the graph structure $\mA$, and more specifically, the neighborhoods $\{N_1, N_2,\ldots, N_{|V|}\}$ of nodes in the graph $G$ (and/or input features $\mX$) are often strongly correlated with the sensitive attribute $\vs$.
Therefore, the goal is often to balance the trade-off between fairness and accuracy.

\subsection{Taxonomy of GNN Fairness Techniques}
Techniques for developing GNNs that are fair and unbiased with respect to the properties above can be categorized as pre-processing, in-training, post-processing, and hybrid.
\begin{enumerate}
    \item \textbf{Pre-processing (Sec.~\ref{sec:gnn-preprocessing})}: Using a pre-processing technique to remove bias or unfairness present in the graph structure $\mA$ or input features $\mX$ before using GNNs.
    \item \textbf{In-training (Sec.~\ref{sec:gnn-intraining})}: By modifying the objective function of GNNs to learn fair and unbiased embeddings during training. This can be the addition of constraints or regularization to the objective function or adding attention weights to GATs that focus on fairness weighting.
    \item \textbf{Post-processing (Sec.~\ref{sec:gnn-postprocessing})}: Using a post-processing technique to remove bias from the resulting embeddings of a GNN model. This can be simply adjusting the embeddings to be independent of the sensitive attribute.
    \item \textbf{Hybrid (Sec.~\ref{sec:gnn-hybrid})}: A hybrid technique combines two or more of the previous techniques to ensure a better and more robust degree of fairness with respect to the sensitive attribute(s). An example of this might be to use a preprocessing technique such as rewiring the graph to ensure exact neighborhood fairness~\cite{chen2022FairNeigh} and then using an in-training technique that adds a fairness constraint to the objective of a GNN model to further ensure fairness of the learned embeddings.
\end{enumerate}
Fairness techniques for GNNs are summarized and categorized according to our proposed taxonomy in Table~\ref{table:qual-and-quant-comparison}.
Notably, we propose a simple and intuitive taxonomy that categorizes fairness techniques for GNNs based on the 
(i) type of input graph supported such as homogeneous, bipartite, or heterogeneous,
(ii) type of unfairness mitigation technique based on whether bias/unfairness mitigation is performed as a pre-processing routine, during training, or as a post-processing technique after learning the embeddings, and the
(iii) graph learning task such as node classification, link prediction, or link classification.

\subsection{Graph Tasks}
There are three fundamental tasks that all practical applications of GNNs leverage, namely, whether the application is edge-based, node-based, or graph-based.
These three general graph machine learning tasks can be formulated as:
\begin{align}
    \vz_i &= f(\vh_i) \qquad\qquad \text{(node-based task)} \\
    \vz_{ij} &= f(\vh_i, \vh_j) \qquad\quad\! \text{(edge-based task)} \\
    \vz_{G} &= f\big( \oplus_{i \in V} \vh_i \big) \qquad\! \text{(graph-based task)}
\end{align}
where $\vz_i$, $\vz_{ij}$, and $\vz_{G}$ are the final embeddings of node $i \in V$, potential edge $(i,j)$, and graph $G$.
An example of a node-based application is node classification, whereas examples of edge-based applications include link prediction and link classification~\cite{rossi2012transforming}.
For graph-based tasks, the most common application is graph classification.

\input{table-summary}

\section{Fairness Evaluation Metrics} \label{sec:metrics}
We now present metrics for evaluating fairness at different fundamental levels.
We categorize these fundamental fairness evaluation metrics into 
graph-level fairness metrics (Sec.~\ref{sec:metrics-graph-level}),
neighborhood-level fairness metrics (Sec.~\ref{sec:metrics-neigh-level}),
embedding-level fairness metrics (Sec.~\ref{sec:metrics-embedding-level}), and 
prediction-level fairness metrics (Sec.~\ref{sec:metrics-pred-level}).

\subsection{Graph-level Fairness Metrics}\label{sec:metrics-graph-level}
We first present metrics for evaluating fairness at the graph-level.
Intuitively, graph-level fairness metrics consider the bias that arises from the graph structure $G$ for a specific sensitive attribute $\vs$.
These metrics are largely based on the notion of homophily that is assumed by the vast majority of graph models.
Homophily is the notion that nodes that are neighbors (adjacent) are more likely to share the same attribute value.
Note that these fairness evaluation metrics are independent of the trained model and its predictions.

One such simple metric for measuring the homophily in a graph is as follows:
\begin{Definition}[\bf Homophily Ratio $h$]
Given a graph $G=(V,E)$ and a sensitive attribute $\vs$ with $|S|$ unique values, then let $\mC \in \RR^{|S| \times |S|}$ be defined as 
\begin{align}
C_{ij} = |\{(u, v): (u, v) \in E \wedge s_u = i \wedge s_v = j\}|
\end{align}
Intuitively, $C_{ij}$ is the frequency that two nodes connected by an edge in $G$ have attribute values of $i \in S$ and $j \in S$.
Then, the homophily ratio $h$ of $G$ is:
\begin{align}
    h(G)    = \frac{\sum_{i} C_{ii}}{\sum_{i}\sum_{j} C_{ij}} 
         =  \frac{\sum_{i} C_{ii}}{|E|} 
\end{align}
where $h\in [0, 1]$. 
At the extremes, a graph with $h=1$ implies that all edges in $G$ connect nodes that have the same sensitive attribute value, and therefore, are highly biased, whereas for $h=0$, then we have the opposite where edges in $G$ connect to nodes with completely different labels.
\end{Definition}

There is also another commonly used metric based on the notion of assortativity:
\begin{Definition}[\bf Assortativity Coefficient $r$]
Given a graph $G=(V,E)$ and a sensitive attribute $\vs$ with $|S|$ unique values, then let 
$\mF \in \RR^{|S| \times |S|}$ be defined as 
\begin{align}
F_{ij} = \frac{|\{(u, v): (u, v) \in E \wedge s_u = i \wedge s_v = j\}|}{|E|}
\end{align}
where $F_{ij}$ is the fraction of edges in $G$ that connect two nodes with attribute values $i \in S$ and $j \in S$.
Notice that $\sum_{i,j} F_{ij} = 1$.
Let $a_i=\sum_{j}F_{ij}$ and $b_j=\sum_{i}F_{ij}$, then the assortativity coefficient $r$ of $G$ is:
\begin{align}
    r(G) = \frac{\sum_{i}F_{ii} - \sum_{i}a_i b_i}{1 - \sum_{i} a_i b_i}
\end{align}
where $r(G) \in [-1,1]$.
Intuitively, $r(G)=1$ implies that all edges in $G$ are between nodes with the same sensitive attribute value whereas $r(G)=0$ implies the opposite, that is, all edges in $G$ are between nodes with different sensitive attribute values.
\end{Definition}

These graph-level metrics are important to understand fairness with respect to only the graph structure and sensitive attributes.
More importantly, suppose a pre-processing fairness approach is used over the initial graph $G$ to make it more fair, thus resulting in another modified graph $G^{\prime}$.
The graph-level fairness evaluation metrics can be used over this new modified graph $G^{\prime}$ to evaluate whether it is more fair or not compared to the original graph or even another graph derived from another approach.
These evaluation metrics can also be used internally during the training process.

\subsection{Neighborhood-level Fairness Metrics}\label{sec:metrics-neigh-level}
We now formally present a neighborhood fairness metric that can be leveraged prior to training a graph neural network model to determine the overall localized fairness in the graph with respect to one or more sensitive attributes.
This metric indicates the impact on fairness from the neighborhoods on the learned embeddings.
In other words, it reveals the overall local fairness when a GNN-based approach is used since these methods all leverage neighborhoods for learning the embeddings of the nodes in the graph.
Therefore, this metric can reveal the overall fairness apriori to training a large-scale GNN model, and based on this, can leverage our approach or future state-of-the-art to mitigate the identified fairness issues that are revealed by the neighborhood fairness metric.
More formally, the entropy-based neighborhood fairness metric is defined as follows:
\begin{Definition}[Local Node Neighborhood Fairness]\label{def:entropy-based-neigh-fairness-metric}
Let $\vc_i$ be the vector of the frequency of the sensitive attribute values of the neighbors $N_i$ of node $i$ such that $c_{ik}=|N_i^{k}|$ where $N_i^{k}$ is the subset of $N_i$ with sensitive attribute value $k$.
Then the neighborhood fairness metric quantifying the localized fairness of a neighborhood of a node $i$ is:
\begin{align}\label{eq:entropy-based-node-neigh-fairness-metric}
    \mathbb{F}(\vp_i) = -\sum_{k} \; p_{ik} \log p_{ik}
\end{align}\noindent
where $\vp_i = \frac{\vc_i}{\sum_k c_{ik}}$ is the probability distribution vector $\sum_k p_{ik}=1$ of node $i$.
Intuitively, when $\mathbb{F}(\vp_i)=1$, then the neighborhood of $i$ is said to be completely fair, as no information is revealed from the neighborhood of $i$ about the sensitive attribute value of $i$.
In other words, when $\mathbb{F}(\vp_i)=1$, the neighborhood $N_i$ leaks no information about the sensitive attribute of $i$ (maximum fairness).
Conversely, when $\mathbb{F}(\vp_i)=0$, then knowing $\vp_i$ reveals significant information about the sensitive attribute (least uncertainty).
Hence, $\mathbb{F}(\vp_i)=0$ indicates a neighborhood with minimum fairness (max unfairness) whereas $\mathbb{F}(\vp_i)=1$ indicates a neighborhood with maximum fairness.
\end{Definition}
Notice that maximum neighborhood fairness is achieved when $\mathbb{F}(\vp_i)=1$, that is, $\vp_i$ is the uniform probability distribution, therefore, revealing no information about the sensitive attribute value $s_i$ of node $i$.
Conversely, maximum neighborhood unfairness is achieved when $\mathbb{F}(\vp_i)=0$, indicating that the sensitive attribute value of $i$ is deterministic, that is, able to be predicted with no uncertainty.
Using Definition~\ref{def:entropy-based-neigh-fairness-metric}, we define the overall neighborhood fairness metric of a graph $G$ is defined as follows:
\begin{Definition}[Neighborhood Fairness]\label{def:neigh-fairness-metric-G}
The neighborhood fairness $\mathbb{F}(G)$ of a graph $G$ is 
\begin{align}\label{eq:neigh-fairness-metric-G}
    \mathbb{F}(G) = \frac{1}{|V|} \sum_{i \in V} \mathbb{F}(\vp_i)
\end{align}\noindent
where $\mathbb{F}(G)$ is an intuitive metric characterizing the inherit fairness of $G$ over all the local neighborhoods. 
Thus, capturing the local fairness of the graph $G$ with respect to the sensitive attribute $\vs = \big[\,s_1\, s_2\, \cdots\, s_i\, \cdots\, s_n\,\big]$.
\end{Definition}
Since neighborhoods lie at the heart of all GNN-based methods~\cite{wu2020comprehensive}, the fairness of the trained GNN models and the resulting node embeddings are fundamentally tied to the neighborhoods used to train these models.

\subsection{Embedding-level Fairness Metrics} \label{sec:metrics-embedding-level}
To measure the fairness of the learned embeddings, one can leverage the notion of representation bias (RB).
This metric enables one to understand if the node embeddings given as output by some arbitrary approach can be leveraged by an adversary to recover the sensitive attribute values of the nodes in the graph.
More formally, for classifier $c$, let $P_c(s,\vz_i)$ denote the estimated probability that node $i$ with embedding $\vz_i$ has sensitive attribute value $s \in S$.  
Then representation bias (RB) score~\cite{pmlr-v119-buyl20a} is:
\begin{align}\label{eq:rep-bias}
    \text{RB} = \sum_{s \in S} \frac{1}{V_s} \texttt{AUC}\big(\{P_c(A(j) | \vz_j) | j \in V_s \}\big)
\end{align}
where $V_s = \{j \in V | A(j) = s \}$ and $A(j)$ is the sensitive attribute value for node $j$.
Eq.~\ref{eq:rep-bias} uses weighted one-vs-rest AUC score to measure prediction performance.
Intuitively, if a model learns fair embeddings, then the classifier trained using the node embeddings should perform poorly (close to random if truly independent).
However, if we are able to predict the sensitive attribute of a node with high accuracy using only the learned embeddings, then they are obviously not independent.

\subsection{Prediction-level Fairness Metrics} \label{sec:metrics-pred-level}

\subsubsection{Statistical Parity (SP)}
The statistical parity (SP) metric (also called demographic parity, or DP) measures the difference between the group-level selection rates of the largest and the smallest groups.
More formally, given the prediction $\hat{Y}$ along with the sensitive attribute value $s$, $\Delta \mathtt{SP}$ is:
\begin{align}\label{eq:SP-metric}
    \Delta \mathtt{SP} = \abs{\mathbb{P}(\hat{Y}=1 \,|\, s=1) \,-\, \mathbb{P}(\hat{Y}=1 \,|\, s=0)}
\end{align}
where $\Delta \mathtt{SP}=0$ implies all groups have the same selection rates, and thus, completely fair.
Statistical parity measures the preferential treatment gap between the groups.
However, $\Delta \mathtt{SP}$ does not consider whether the individual is qualified or not since it does not consider the ground-truth $Y$.
Note that Eq.~\ref{eq:SP-metric} is defined for sensitive attributes with only two groups, though, is easy to generalize to $k$ groups by considering the difference between the largest and smallest group-level selection rate across all values of the sensitive attribute:
\begin{align}\label{eq:SP-metric-max}
    \Delta \mathtt{SP} = \abs{\max_{s} \mathbb{E}\big[\hat{Y} \,|\, S=s\big] \,-\, \min_{s} \mathbb{E}\big[\hat{Y} \,|\, S=s\big]}
\end{align}

The above formulation also generalizes to the link prediction task. 
More formally, statistical parity for a link prediction function $h : V \times V \rightarrow \{0,1\}$ is:
\begin{align}\label{eq:SP-metric-LP}
    \Delta \mathtt{SP} &= \abs{\mathbb{P}\big(h(v,u)=1 \,|\, s_v\not=s_u\big) \,-\, \mathbb{P}\big(h(v,u)=1 \,|\, s_v=s_u\big)}\\
    &= \abs{\mathbb{P}\big(h(v,u)=1 \,|\, s_v \oplus s_u = 1\big) \,-\, \mathbb{P}\big(h(v,u)=1 \,|\, s_v \oplus s_u = 0\big)}\nonumber
\end{align}
where $s_v \oplus s_u = 1$ (or $s_v\not=s_u$) implies that $v$ and $u$ belong to different groups whereas $s_v \oplus s_u = 0$ (or $s_v=s_u$) implies they belong to the same group.
Hence, in the case of link prediction, we only consider whether the sensitive attribute values are the same $s_v=s_u$ or not $s_v\not=s_u$, since an edge either exists or not, $h(v,u) \in \{0,1\}$.

\subsubsection{Equal Opportunity (EO)}
The equal opportunity metric requires the non-discrimination only within the ``advantaged'' outcome group.
More formally, given the ground-truth $Y$, the prediction $\hat{Y}$, along with the sensitive attribute value $s$, $\Delta \mathtt{EO}$ is:
\begin{align}\label{eq:EO-metric}
    \Delta \mathtt{EO} = \abs{\mathbb{P}(\hat{Y}=1 \,|\, Y=1,\, s=1) \,-\, \mathbb{P}(\hat{Y}=1 \,|\, Y=1,\, s=0)}
\end{align}
where lower values of $\Delta \mathtt{EO}$ imply better fairness.
The above equal opportunity metric is a relaxation of the equalized odds metric~\cite{hardt2016equality} that measures the difference of true positives, true negatives, false positives and false negatives between the groups.
More formally, given the ground-truth $Y$, the prediction $\hat{Y}$, along with the sensitive attribute value $s$, $\Delta \mathtt{EO}$ is:
\begin{align}\label{eq:equalized-odds-metric}
    \Delta \mathtt{EO} = \abs{\mathbb{P}(\hat{Y}=1 \,|\, Y=y,\, s=1) \,-\, \mathbb{P}(\hat{Y}=1 \,|\, Y=y,\, s=0)},\quad y\in\{0,1\}
\end{align}
where $\Delta \mathtt{EO}=0$ implies all groups have the same true positive, true negative, false positive, and false negative rates, and are therefore fair.

\section{GNN Fairness Techniques} \label{sec:gnn-fairness-techniques}
For graph fairness, techniques can generally take one of three entry points for their mitigation -- modifying graphs before training with pre-processing (Sec.~\ref{sec:gnn-preprocessing}), modifying the training process (Sec.~\ref{sec:gnn-intraining}), modifying the outputs with post-processing (Sec.~\ref{sec:gnn-postprocessing}), or a hybrid approach that combines two or more mitigation techniques at different stages (Sec.~\ref{sec:gnn-hybrid}).
We summarize and categorize GNN fairness techniques in Table~\ref{table:qual-and-quant-comparison} using the proposed taxonomy that categorizes 
fairness techniques for GNNs based on the 
(i) type of input graph supported (\eg, homogeneous, bipartite, heterogeneous),
(ii) type of unfairness mitigation technique based on whether bias mitigation is performed during pre-processing, training, or post-processing, and 
(iii) graph learning task such as node classification, link prediction, or link classification.

\begin{figure}
\centering
\includegraphics[width=0.9\linewidth]{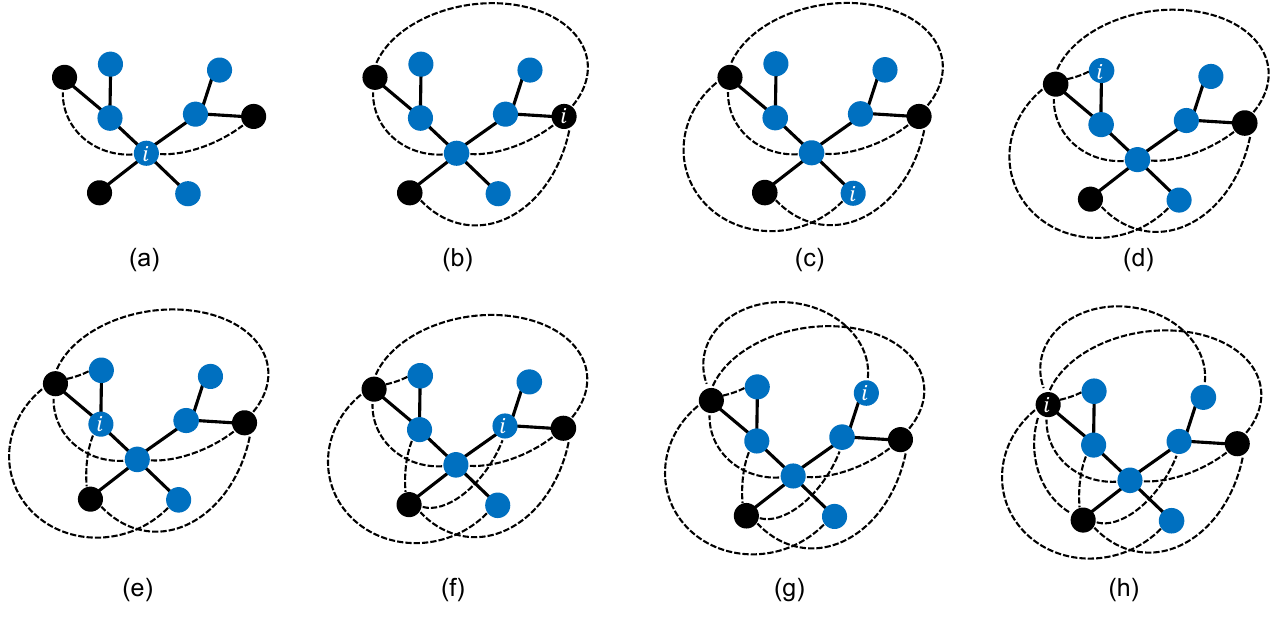}
\caption{Exact neighborhood fairness is NP-hard for GNNs even at the graph-level. 
Greedy fairness optimization via neighborhood edge augmentation.
In each step, a vertex $i \in V$ is selected, and the neighborhood is modified to make it fair, \eg, by adding two edges as shown in (a).
As this process continues for every node in a greedy fashion, there is no guarantee that the subsequent nodes remain fair, unless we explicitly incorporate this constraint, which makes this problem in general very complex, and even with this additional constraint to revisit nodes or carefully change the graph to ensure nodes visited previously remain fair, there is no guarantee that all such nodes can actually be fair.
This intuition is also useful for understanding in-training methods discussed in Sec~\ref{sec:gnn-intraining} that sample or augment neighborhoods to reduce bias for each node during the neighborhood aggregation process when training the GNN.  
}
\label{fig:greedy-opt-edge-aug}
\end{figure}

\subsection{Pre-Processing} \label{sec:gnn-preprocessing}
Pre-processing techniques remove bias or unfairness before GNN training occurs, by targeting the input graph structure $\mA$, input features $\mX$, or both.
For instance, work by~\citet{FairDrop} proposed a pre-processing approach that  randomly removes edges from the graph before training to debias the resulting GNN model. 
More recently,~\citet{chen2022FairNeigh} developed a GNN fairness framework based on the proposed notion of neighborhood fairness. The framework consists of two main components where the first component constructs unbiased and fair neighborhoods by adding and removing edges to ensure each neighborhood is unbiased with respect to the sensitive attribute while preserving structures important for prediction tasks such as link prediction and classification. The second component provides additional flexibility by enabling the fair neighborhoods to be modified via a function to capture certain application or data-dependent constraints. These fair neighborhoods are then leveraged by any arbitrary GNN model to learn fair embeddings for downstream graph learning tasks.
An intuitive illustration showing the difficulty of guaranteeing exact fairness with respect to the neighborhoods is shown in Figure~\ref{fig:greedy-opt-edge-aug}. In particular, we see that even in this simple example, it becomes impossible to ensure fairness with respect to each neighborhood in the graph, since when one neighborhood is made fair, it can impact the surrounding neighborhoods. 
It is also straightforward to see that an iterative optimization approach to solve this would require a significant computational cost without any guarantees of  fairness across each neighborhood, and the neighborhoods that have the largest impact are often the ones that have the most impact when training GNNs, since they are the ones that connect to many other nodes, and therefore, updating the neighborhood, and even the embedding when using this approach during training of the GNN impacts the embeddings of all other nodes connected.
\citet{current2022fairmod} studied a few graph modification strategies that perform either microscopic or macroscopic edits to the input graph.
One of the proposed methods adds a new node for each existing node to balance biases in the graph, whereas the other methods only include a fixed number of existing nodes and include weights for the edges as a means to debias the graph for GNN training.
Another approach called FairAdj~\cite{FairAdj} seeks to learn a fair adjacency matrix for a downstream link prediction task by updating the normalized adjacency matrix while keeping the original graph unchanged. 
This approach rewires the graph to preserve structural constraints for fairness while trying to also preserve accuracy as much as possible.
Furthermore, they introduce dyadic fairness that expects the prediction of a link between two nodes to be statistically independent of their sensitive attributes, hence, $P(g(u,v) | S(u)=S(v)) \!=\! P(g(u,v) | S(u)\not=S(v))$.
\citet{yang2022obtaining} proposed data reparation through optimal transport techniques to obtain dyadic fairness. 
Similarly,~\citet{laclau2021all} proposed a repairing procedure for the graph adjacency matrix with a trade-off between group and individual fairness.

\subsection{In-Training} \label{sec:gnn-intraining}
Most GNN-based bias mitigation techniques have focused on modifying the objective function of GNNs to learn fair and unbiased embeddings during training. This can be through the addition of constraints or regularization to the objective function, adding attention weights to GATs that focus on fairness weighting, or careful sampling of the explicit neighborhoods for updating the embedding via aggregation, as well as many other ways to mitigate bias during training, which are discussed in detail below.
For instance, DegFairGCN~\cite{liu2023generalized} considers two groups of nodes based on low and high-degree when performing neighbor aggregations, namely, $\mathcal{S}_0=\{v \in \mathcal{V} | \mathrm{deg}_1(v) \leq K\}$ and $\mathcal{S}_1 = \mathcal{V} \setminus \mathcal{S}_0$ where $\mathcal{S}_0$ is the low-degree nodes, $\mathcal{S}_1$ is the high-degree node group, and $K$ is a threshold for creating such groups.
Using these two groups, they modify the neighborhood aggregation used to consider these two groups differently, attempting to debias them accordingly during training.
Instead of using traditional sensitive attributes for fairness evaluation, that work used node degree as the sensitive attribute, which is problematic.
FairEdit leverages both greedy edge additions and deletions to improve fairness in GNNs~\cite{loveland2022fairedit}.
Recent work by~\citet{kose2023fairgat} developed an attention mechanism that mitigates bias called FairGAT.
The fairness-aware attention mechanism can be leveraged in other attention-oriented GNNs as well~\cite{lee2019attention}.
All these approaches discussed thus far have focused on reducing bias in the neighborhood used for aggregation either by sampling, modifying, or reweighting the nodes in the neighborhood right before its used for aggregation during training.
However, performing aggregation using these locally ``fair'' neighborhoods has even fewer guarantees than approaches that modify the graph structure before training, see Fig~\ref{fig:greedy-opt-edge-aug} and the discussion in Sec.~\ref{sec:gnn-preprocessing} for intuition.
Other work by~\citet{palowitch2019monet} introduced a neural network architecture component called MONET that performs training-time debiasing by ensuring the embeddings are trained on a hyperplane orthogonal to the metadata. 
\citet{agarwal2021towards} developed a novel objective function to account for fairness and stability called NIFTY. They also introduce a layer-wise weight normalization to enforce fairness in the GNN architecture.
Further, \citet{pmlr-v119-buyl20a} proposed a Bayesian method called DeBayes that leverages a biased prior to learn debiased embeddings.
\citet{dong2021individual} introduced a rank-based approach called REDRESS for mitigating individual unfairness in GNNs where the goal is to ensure GNNs infer similar predictions for individual nodes that are similar to one another.
The approach jointly optimizes the utility maximization of GNNs and rank-based individual fairness in an end-to-end fashion.

\citet{zhu2023fairness} proposed a fairness-aware message-passing framework for GNNs called GMMD for node classification that seeks to jointly optimize both representation fairness and graph smoothing.
Similarly, \citet{lin2023bemap} developed a balanced message-passing approach for GNNs called BeMap. This approach uses a sampling strategy to balance the number of 1-hop neighbors of each type for every node in the graph. This is in principle similar to the first step of FairNeigh, though performed during training.
\citet{dong2022edits} developed an approach called Edits for mitigating both attribute-based bias as well as structural bias in GNNs based on the Wasserstein distance. However, attribute and structural debiasing are independent of one another, as opposed to being jointly considered, which is important since GNNs are trained by leveraging both.
More recently, \citet{he2023efficient} proposed an efficient approach called FairMILE for ensuring fairness in GNNs via a multi-level framework that leverages graph coarsening to obtain base embeddings and then refines these to obtain an embedding for each node of the graph.
There are also many other in-training approaches that leverage an adversarial framework, by incorporating the objective that an adversarial model should not have high accuracy in predicting the sensitive attribute~\cite{wang2022towards,liu2022dual,khajehnejad2020adversarial,singh2021temporal}.

There have also been a number of works focused on GNN-based recommendation~\cite{xu2023fairness,wu2022equipping,salganik2022analyzing,medda2023gnnuers,liu2022dual,chizari2023bias,wu2023faster}.
FairRec~\cite{patro2020fairrec} was proposed for the closely related task of recommendation.
In particular, they studied fairness in recommender systems involving customers and producers, and proposed an approach called FairRec that is based on fair allocation of indivisible goods.
FairRec guarantees at least maximin share of exposure for most producers and envy-free up to one good fairness for every customer.
\citet{li2021towards} proposed an adversarial in-training method to learn fair user embeddings for fair recommendations.
Separately,~\citet{li2022fairsr} designed a framework for fair sequential recommendations, which can both do end-to-end training and also learn fairness-aware preference graph embeddings.
There have also been some recent works that exploit communities to obtain fair link predictions in complex networks, such as HM-EIICT~\cite{saxena2021hm}.
\citet{tsioutsiouliklis2021fairness} developed algorithms for fairness in the PageRank algorithm, requiring fairness on the proportion of PageRank score assigned to each group, and then fairness on the derived personalized PageRank to each node.

Recent work has also considered mitigating fairness issues in a wide range of different types of graphs, including hypergraphs~\cite{weihypergcl}, heterogeneous information networks~\cite{cao2023fairhelp,zeng2021fair}, 
knowledge graphs~\cite{wang2022towards,vannur2021data}, and even temporal networks~\cite{song2022individual,xu2021highair}.
For hypergraphs, \citet{weihypergcl} proposed HyperGCL and show that their method for augmenting hypergraphs improves fairness in representation learning.
\citet{cao2023fairhelp} proposed FairHELP for deriving fair embeddings for heterogeneous information networks. 
Most recently, temporal networks have been considered by~\citet{song2022individual} where they propose an approach for improving individual fairness for dynamic GNNs such as EvolveGCN.
For this, they introduce a simple regularization-based method to achieve individual fairness in the dynamic graph setting for GNNs.
Other papers have identified node degree as a source of bias~\cite{tang2020investigating, kang2022rawlsgcn, jiang2022fmp}. The latter defines node-degree disparity in terms of the Rawlsian difference principle, and proposes a RawlsGCN-Graph pre-processing method and a RawlsGCN-Grad in-processing method for fair predictive accuracy. 
Recent work by~\citet{loveland2022graph} investigated fairness in GNNs when neighborhoods in the graph are heterophilous as opposed to homophilous. In this setting, they find that several fairness metrics can be significantly improved when leveraging heterophilous GNNs that naturally handle disassortativity.

\subsection{Post-Processing} \label{sec:gnn-postprocessing}
Post-processing techniques take the output embeddings of a GNN model and remove bias from them. Techniques include adding filters or otherwise removing information about the sensitive attribute from those embeddings.
\citet{masrour2020bursting} addressed the problem of link prediction homophily with postprocessing, as well as an adversarial framework.
\citet{fisher2020debiasing} developed techniques for debiasing knowledge graph embeddings.
\citet{dai2021say} proposed FairGNN, a framework for fair node classification using GNNs given limited sensitive attribute information. 
\citet{bose2019compositional} investigated incorporating compositional fairness constraints into graph embeddings, creating sensitive attribute filters that can be optionally applied after training.
FairGO~\cite{wu2021learning} was proposed to ensure fairness by taking the original embeddings from a method and then leveraging a composition of filters that transform the embeddings to a new filtered embedding space to improve fairness. This transformation leverages adversarial learning of a user-centric graph to obfuscate the sensitive attribute from the underlying embeddings.
\citet{kose2023fairness} designed a fairness-aware filter to reduce the bias in the learned embeddings from GNNs by essentially removing the sensitive information. This technique can be used in other GNN designs. They also demonstrate the effectiveness theoretically compared to the fairness-agnostic embedding that arises without their fairness-aware filter.

\subsection{Hybrid}\label{sec:gnn-hybrid}
A hybrid technique combines two or more techniques that are used at different stages (\eg, pre-processing, during training, and post-processing) to ensure a better and more robust degree of fairness with respect to the sensitive attribute(s). 
Few papers explicitly propose hybrid methods, which combine two or more of the previous techniques, but it is very much a possibility to improve fairness and robustness through some combinations. An example of this might be to use a preprocessing technique such as rewiring the graph to ensure exact neighborhood fairness~\cite{chen2022FairNeigh} and then using an in-training technique that adds a fairness constraint to the objective of a GNN model to further ensure fairness of the learned embeddings.
While there have not been many hybrid fairness techniques for GNNs, there has been one work by \citet{inform2020} that focused on a related graph learning method for graph mining tasks. In particular, InFoRM first modifies the graph structure to remove bias, then attempts to debias the mining model by solving an optimization problem, and finally solves a similar problem for debiasing the results from the mining model.
It is straightforward to leverage many of the GNN fairness techniques discussed in the previous sections together to obtain novel hybrid approaches for providing even more fair GNNs and results with potentially stronger fairness guarantees as well.

\input{table-datasets}

\section{Datasets}
We summarize datasets commonly used for evaluating fairness of GNNs in Table~\ref{table:datasets}.
Notably, the datasets are organized by application domain including 
\textcolor{googlered}{\textbf{recommendation (bipartite graphs)}}, 
\textcolor{googleblue}{\textbf{social networks}},
\textcolor{googlegreen}{\textbf{collaboration networks}},
\textcolor{googlepurple}{\textbf{web graphs}},
\textcolor{darkblue}{\textbf{similarity graphs}}, and
\textcolor{lightyellow}{\textbf{citation networks}}.
Furthermore, $|V|$ and $|E|$ are the number of nodes and edges, whereas $|S|$ is the number of unique values of the sensitive attribute $S$.
We also denote $|\mathcal{S}|$ as the number of sensitive attributes and $|\mX|$ as the number of input features.
While most work focuses on a single sensitive attribute, there are some datasets with multiple sensitive attributes that can be used for fairness techniques designed for multiple sensitive attributes.
Furthermore, we also summarize the various fair graph learning tasks that each dataset was used, which include link prediction (LP), node classification (NC), and link classification (LC).
Notably, there is only a single link classification dataset used in the literature. This may be due to the task being close to real-world data found in industry, but also evaluating fairness in link classification requires having one or more sensitive attributes on the nodes or links.

\section{Open Problems \& Challenges}\label{sec:open-problems-and-challenges}
In this section, we discuss open problems and highlight important challenges for future work.

\subsection{Feature vs. Feature-less Setting}
Previous work on developing fairness techniques for GNNs has mainly focused on graphs with features. The reason for this is quite simple. 
GNNs require an initial feature matrix, and therefore, most work has simply used datasets that naturally come with an input feature matrix.
However, this severely limits the graph datasets used for evaluation to those that have one or more sensitive attributes, as well as an entire feature matrix.
Most importantly though, input features are often highly correlated with the sensitive attribute, and therefore potentially (and often) add multiple confounding factors when evaluating fairness techniques that are mostly focused on the structure of the graph, as opposed to the correlation of input features with respect to the sensitive attribute. 

For these fundamental reasons, one should also consider a new setting when evaluating techniques for improving fairness in GNNs, which we call the \emph{feature-less setting}. 
In this proposed setting, we do not include any underlying features as input, and instead initialize the feature matrix either uniformly at random or based on the graph structure, e.g., using SVD or an unsupervised embedding approach such as node2vec~\cite{https://doi.org/10.48550/arxiv.1607.00653} or DeepGL~\cite{rossi2018deep}.
This new feature-less setting may actually be more useful, as most graphs do not naturally come with input features (see NetworkRepository by~\citet{nr}), and therefore, it opens the door for evaluation on a much larger scale and gives rise to entirely new use cases and practical applications for such approaches.
Nevertheless, one can also study graphs with features under this setting by simply ignoring them.
Studying both the feature and feature-less setting allows for a better evaluation and understanding of the approach under different conditions and controlling for different factors that may influence the fairness, accuracy, and the underlying conclusions that are drawn from the experiments.
Understanding how previous work performs in this setting remains an open problem.

\subsection{Theoretical Limits}
Understanding the theoretical limits in terms of the fairness and accuracy trade-offs and deriving theoretical guarantees for such techniques is fundamentally important.
Despite this, theoretically analyzing existing fairness techniques for GNNs remains a largely open problem for future work.

\subsection{Link Classification}
While most work has focused on developing techniques for either node classification or link prediction, the problem of link classification remains largely unexplored.
In this task, we are given a small fraction of link labels for training and need to predict the remaining held-out labels of the links.
This is fundamentally different from link prediction since in link classification we are given the entire graph $G$ along with a sensitive attribute on the nodes that is never seen by the algorithm, and need to correctly infer the label of the link (which is already existing in the graph).
Link classification in GNNs are often a multi-class problem where the number of unique labels to infer is much larger than inferring only two labels, which is the simplest binary link classification task.

\subsection{Heterogeneous and Temporal Networks}
Most work has only focused on developing fairness techniques for GNNs on simple graphs. However, it is unclear how such techniques perform when the graph is heterogeneous, that is, nodes and edges may be of completely different types.
Similarly, ensuring fairness when the graph structure and possibly even its attributes of it are changing over time remains an open and challenging problem.
These different types of networks may also lead to the need for fairness metrics for the evaluation of such techniques for these important types of networks.

\subsection{Large Language Models}
Recently, GNNs have found applications in language models~\cite{meng2021gnn}.
Fairness in such models is of fundamental importance due to their wide-scale use in many applications, yet very challenging due to how these models are trained.
Future work should focus on developing fair and unbiased GNN-based language models.

\section{Conclusion} \label{sec:conc}
Given the importance of GNNs due to their representational power and state-of-the-art predictive performance, this paper has surveyed techniques for improving the fairness of GNNs.
After presenting a taxonomy for fairness in GNNs that categorizes techniques based on the 
type of input graph supported, 
type of fairness technique (post-processing, in-training, post-processing), and the 
graph learning task.
We also introduce an intuitive taxonomy for graph fairness evaluation metrics including graph-level fairness, neighborhood-level fairness, embedding fairness, and prediction-level fairness metrics for GNNs.
Furthermore, we summarize the graph datasets useful for benchmarking GNN fairness techniques and categorize them according to the domain and graph learning task.
As discussed in Section~\ref{sec:open-problems-and-challenges}, there remains significant work to do.
One important and largely open problem is understanding the theoretical limits in terms of the fairness and accuracy trade-offs and deriving theoretical guarantees for such techniques.
Theoretically analyzing existing fairness techniques across the different categories of approaches (pre-processing, in-training, and post-processing) remains a largely open problem for future work as well.

\bibliographystyle{ACM-Reference-Format}
\bibliography{paper}

\end{document}

%% file: preamble.tex
\usepackage{xcolor}
\definecolor{thedarkblue}{RGB}{0,0,120} 
\definecolor{mydarkblue}{rgb}{0,0.08,0.45} 
\definecolor{darkblue}{rgb}{0,0.08,180}
\colorlet{TufteRed}{red!80!black}
\definecolor{theblue}{RGB}{0,0,180}
\colorlet{thered}{TufteRed}
      
\usepackage{microtype}
\usepackage{balance}

\usepackage{booktabs}
\usepackage{tabularx}

\usepackage{amsmath,amsthm}

\usepackage{comment}

\usepackage{tikz}
\usepackage{verbatim}
\usetikzlibrary{arrows}
\usetikzlibrary{shapes,snakes}
\usetikzlibrary{decorations.pathmorphing} 
\usetikzlibrary{fit}			
\usetikzlibrary{backgrounds}	

\usepackage{ragged2e}
\usepackage{multirow}
\usepackage{microtype}
\usepackage{balance}
\usepackage{setspace}

\graphicspath{{./}{./graphics/}}
\newcolumntype{H}{>{\setbox0=\hbox\bgroup}c<{\egroup}@{}}

\newcolumntype{R}[1]{>{\RaggedLeft\arraybackslash}} 
\newcolumntype{L}[1]{>{\RaggedRight\arraybackslash}} 

\newcommand\TTT{\rule{0pt}{3.2ex}}

\newcommand{\abs}[1]{\left|#1\right|}

\newcommand{\eg}{\emph{e.g.}}

\newtheorem{Definition}{\hspace{-1em}\bfseries{Definition}}

\AtBeginEnvironment{pmatrix}{\setlength{\arraycolsep}{2pt}}

\providecommand{\mat}[1]{\boldsymbol{\mathrm{#1}}}%
\renewcommand{\vec}[1]{\boldsymbol{\mathrm{#1}}}

\DeclareMathOperator{\hugeE}{\mbox{\huge\raise-0.3ex\hbox{E}}}
\DeclareMathOperator{\p}{\mathbb{P}}
\DeclareMathOperator{\hugep}{\mbox{\huge\raise-0.3ex\hbox{$\p$}}}

\newcommand{\RR}{\mathbb{R}}

\providecommand{\mA}{\ensuremath{\mat{A}}}

\providecommand{\mC}{\ensuremath{\mat{C}}}

\providecommand{\mF}{\ensuremath{\mat{F}}}

\providecommand{\mH}{\ensuremath{\mat{H}}}

\providecommand{\mX}{\ensuremath{\mat{X}}}

\providecommand{\vc}{\ensuremath{\vec{c}}}

\providecommand{\vh}{\ensuremath{\vec{h}}}

\providecommand{\vp}{\ensuremath{\vec{p}}}

\providecommand{\vs}{\ensuremath{\vec{s}}}

\providecommand{\vx}{\ensuremath{\vec{x}}}

\providecommand{\vz}{\ensuremath{\vec{z}}}

%% file: table-summary.tex
\newcommand\TE{\rule{0pt}{2.0ex}}
\newcommand\BE{\rule[-1.1ex]{0pt}{0pt}}

{
\newcolumntype{C}{ >{\centering\arraybackslash} m{4cm} } 
\providecommand{\rotateDeg}{90}
\setlength{\tabcolsep}{1.2pt}
\providecommand{\rotDeg}{70}
\definecolor{verylightgreennew}{RGB}	{220,255,220}
\definecolor{verylightrednew}{RGB}		{255, 230, 230}
\definecolor{verylightreddarker}{HTML} {FFCBCB} 
\definecolor{verylightrednew}{RGB}		{255, 230, 230}
\definecolor{verylightrednewlighter}{RGB}		{255, 229, 239}
\definecolor{lightgraynew}{rgb}{0.9,0.9,0.9}
\definecolor{newgray}{RGB}{0.7,0.7,0.7}
\providecommand{\cellsz}{0.34cm} 
\providecommand{\cellszlg}{0.36cm} 
\renewcommand{\cm}{{\color{greencm}\normalsize\cmark}}
\renewcommand{\cmgray}{{\color{newgray}\normalsize\cmark}}
\renewcommand{\xm}{{\color{verylightreddarker}\normalsize\xmark}}
\newcommand\BBBBB{\rule[1.6ex]{0pt}{1.6ex}}
\newcommand\BBBBBB{\rule[-1.1ex]{0pt}{0pt}} 
\newcommand{\sysName}[1]{{\sf
\BBBBBB
#1
}}
\providecommand{\cellsomewhat}{
\BBBBB
\cmgray
\cellcolor{lightgraynew}
}
\providecommand{\cellno}{
\BBBBB
\xm
\cellcolor{verylightrednew}}
\providecommand{\cellyes}{
\BBBBB
\cm
\cellcolor{verylightgreennew}
}

\newcolumntype{H}{>{\setbox0=\hbox\bgroup}c<{\egroup}@{}} 
\begin{table}[t!]
\def\arraystretch{1.2} 
\scriptsize
\caption{%
Overview of the proposed taxonomy for fairness-aware GNNs.
Techniques are categorized by the type of input graph supported, whether the approach focuses on pre-processing, in-training, or post-processing, as well as the task the technique was designed.
Using this taxonomy, we provide a qualitative and quantitative comparison of fairness methods for graphs.
}
\label{table:qual-and-quant-comparison}
\begin{minipage}{\columnwidth}
{\begin{center}
\begin{tabular}
{P{2mm}
l
c 
!{\vrule width 0.8pt} 
P{\cellsz} P{\cellsz} P{\cellsz} 
P{\cellsz} P{\cellsz} 
P{\cellsz} 
H 
HHH
H@{}
!{\vrule width 0.6pt}
P{\cellszlg} 
P{\cellszlg} 
P{\cellszlg} 
H
@{}
!{\vrule width 0.6pt} 
P{\cellsz} P{\cellsz} P{\cellsz} P{\cellsz} 
H
P{\cellsz} 
!{\vrule width 0.8pt}
@{}
}

& & 
& \multicolumn{7}{c}{\textsc{\bfseries\scshape Input}}
&&&&
& \multicolumn{4}{c!{\vrule width 0.6pt}}{\textsc{\bfseries\scshape Approach}}
& \multicolumn{6}{c!{\vrule width 0.6pt}}{\textsc{\bfseries\scshape Task}} 
\\

& \multicolumn{1}{l}{\textbf{\bfseries\scshape \small Method}} & 
& 
\rotatebox{\rotateDeg}{\textbf{Homogeneous graph}} & 
\rotatebox{\rotateDeg}{\textbf{Bipartite graph}} &
\rotatebox{\rotateDeg}{\textbf{Heterogeneous graph}} &
\rotatebox{\rotateDeg}{\textbf{Knowledge graph}} &
\rotatebox{\rotateDeg}{\textbf{Hypergraph}} &
\rotatebox{\rotateDeg}{\textbf{Temporal graph}} & & 
&&&&
\rotatebox{\rotateDeg}{\textbf{Pre-processing}} &
\rotatebox{\rotateDeg}{\textbf{In-training}} &  
\rotatebox{\rotateDeg}{\textbf{Post-processing}} & 
\rotatebox{\rotateDeg}{\textbf{Other}} & 
\rotatebox{\rotateDeg}{\textbf{Link prediction}} &
\rotatebox{\rotateDeg}{\textbf{Link classification}} &
\rotatebox{\rotateDeg}{\textbf{Node classification}} &
\rotatebox{\rotateDeg}{\textbf{Clustering}} &&
\rotatebox{\rotateDeg}{\textbf{Other}} 
\\
\noalign{\hrule height 0.8pt}

& \sysName{$\mathsf{\sf \bf FairDrop}$}~\cite{FairDrop}
&
& \cellyes
& \cellno
& \cellno
& \cellno 
& \cellno 
& \cellno
& \cellno
& \cellyes
& \cellyes
& \cellyes 
& \cellno
& \cellyes 
& \cellno 
& \cellno
& \cellno
& \cellyes
& \cellno
& \cellno
& \cellno 
& \cellno
& \cellno 
\\

\hline
& \sysName{$\mathsf{\sf \bf FairOT}$}~\cite{laclau2021all}
&
& \cellyes 
& \cellno
& \cellno
& \cellno 
& \cellno 
& \cellno
& \cellno 
& \cellyes  
& \cellno
& \cellyes
& \cellno
& \cellyes
& \cellno
& \cellno 
& \cellno 
& \cellyes
& \cellno
& \cellno
& \cellno 
& \cellno
& \cellno 
\\

\hline
& \sysName{$\mathsf{\sf \bf UGE}$}~\cite{wang2022unbiased}
&
& \cellyes 
& \cellno
& \cellno
& \cellno 
& \cellno 
& \cellno
& \cellno 
& \cellyes  
& \cellno
& \cellyes
& \cellno
& \cellyes
& \cellno
& \cellno 
& \cellno 
& \cellyes
& \cellno
& \cellno
& \cellno 
& \cellno
& \cellno 
\\

\hline
& \sysName{$\mathsf{\sf \bf FairNeigh}$}~\cite{chen2022FairNeigh}
&
& \cellyes
& \cellno
& \cellno
& \cellno 
& \cellno 
& \cellno
& \cellno 
& \cellyes
& \cellyes
& \cellyes 
& \cellno
& \cellyes 
& \cellno 
& \cellno
& \cellno
& \cellyes
& \cellyes
& \cellno
& \cellno 
& \cellno
& \cellno 
\\

\hline
& \sysName{$\mathsf{\sf \bf FairAdj}$}~\cite{FairAdj}
&
& \cellyes 
& \cellno
& \cellno
& \cellno 
& \cellno 
& \cellno
& \cellno 
& \cellyes  
& \cellno
& \cellyes
& \cellno
& \cellno
& \cellyes
& \cellno 
& \cellno 
& \cellyes
& \cellno
& \cellno
& \cellno 
& \cellno
& \cellno 
\\

\hline
& \sysName{$\mathsf{\sf \bf FairRec}$}~\cite{patro2020fairrec}
&
& \cellno
& \cellyes
& \cellno
& \cellno 
& \cellno
& \cellno 
& \cellno 
& \cellno
& \cellno
& \cellno
& \cellno
& \cellno
& \cellyes
& \cellno 
& \cellno 
& \cellyes 
& \cellno
& \cellno
& \cellno 
& \cellno
& \cellno 
\\

\hline
& \sysName{$\mathsf{\sf \bf FairGO}$}~\cite{wu2021learning}
&
& \cellno
& \cellyes
& \cellno
& \cellno 
& \cellno
& \cellno 
& \cellno 
& \cellno
& \cellno
& \cellno
& \cellno
& \cellno
& \cellno
& \cellyes 
& \cellno 
& \cellyes 
& \cellno
& \cellno
& \cellno 
& \cellno
& \cellno 
\\

\hline
& \sysName{$\mathsf{\sf \bf DeBayes}$}~\cite{pmlr-v119-buyl20a}
&
& \cellyes
& \cellyes
& \cellno
& \cellno 
& \cellno
& \cellno 
& \cellno 
& \cellno
& \cellyes
& \cellyes
& \cellno
& \cellno
& \cellyes
& \cellno 
& \cellno 
& \cellyes
& \cellno
& \cellno
& \cellno 
& \cellno
& \cellno 
\\

\hline
& \sysName{$\mathsf{\sf \bf FIPR}$}~\cite{buyl2021kl}
&
& \cellyes 
& \cellyes
& \cellno
& \cellno 
& \cellno
& \cellno 
& \cellno 
& \cellno
& \cellno
& \cellno
& \cellno
& \cellno
& \cellyes
& \cellno 
& \cellno 
& \cellyes
& \cellno
& \cellno
& \cellno 
& \cellno
& \cellno 
\\

\hline
& \sysName{$\mathsf{\sf \bf FairGAT}$}~\cite{kose2023fairgat}
&
& \cellyes
& \cellno
& \cellno
& \cellno 
& \cellno
& \cellno 
& \cellno 
& \cellno
& \cellno
& \cellno
& \cellno
& \cellno
& \cellyes
& \cellno 
& \cellno 
& \cellyes
& \cellno
& \cellyes
& \cellno 
& \cellno
& \cellno 
\\

\hline
& \sysName{$\mathsf{\sf \bf Fairwalk}$}~\cite{rahman2019fairwalk}
&
& \cellyes
& \cellno
& \cellno
& \cellno 
& \cellno
& \cellno 
& \cellno 
& \cellyes
& \cellno
& \cellyes
& \cellno
& \cellno
& \cellyes
& \cellno 
& \cellno 
& \cellyes
& \cellno
& \cellno
& \cellno 
& \cellno
& \cellno 
\\
\hline

& \sysName{$\mathsf{\sf \bf FairEdit}$}~\cite{loveland2022fairedit}
&
& \cellyes
& \cellno
& \cellno
& \cellno 
& \cellno
& \cellno 
& \cellno 
& \cellyes
& \cellno
& \cellno
& \cellno
& \cellno
& \cellyes
& \cellno 
& \cellno 
& \cellno
& \cellno
& \cellyes
& \cellno 
& \cellno
& \cellno 
\\
\hline

& \sysName{$\mathsf{\sf \bf FairEGM}$}~\cite{current2022fairmod}
&
& \cellyes
& \cellno
& \cellno
& \cellno 
& \cellno
& \cellno 
& \cellno 
& \cellno
& \cellyes 
& \cellno
& \cellno
& \cellyes
& \cellno
& \cellno 
& \cellno 
& \cellyes
& \cellno
& \cellno
& \cellno 
& \cellno
& \cellno 
\\
\hline

& \sysName{$\mathsf{\sf \bf HM\text{-}EIICT}$}~\cite{saxena2021hm}
&
& \cellyes 
& \cellno
& \cellno
& \cellno 
& \cellno
& \cellno 
& \cellno 
& \cellno
& \cellno
& \cellyes
& \cellno
& \cellno
& \cellyes
& \cellno 
& \cellno 
& \cellyes
& \cellno
& \cellno
& \cellyes
& \cellno
& \cellno 
\\
\hline

& \sysName{$\mathsf{\sf \bf DegFairGCN}$}~\cite{liu2023generalized}
&
& \cellyes 
& \cellno
& \cellno
& \cellno 
& \cellno
& \cellno 
& \cellno 
& \cellno
& \cellno
& \cellyes
& \cellno
& \cellno
& \cellyes
& \cellno 
& \cellno 
& \cellno
& \cellno
& \cellyes
& \cellno 
& \cellno
& \cellno 
\\
\hline

& \sysName{$\mathsf{\sf \bf GMMD}$}~\cite{zhu2023fairness}
&
& \cellyes 
& \cellno
& \cellno
& \cellno 
& \cellno
& \cellno 
& \cellno 
& \cellno
& \cellno
& \cellyes
& \cellno
& \cellno
& \cellyes
& \cellno 
& \cellno 
& \cellno
& \cellno
& \cellyes
& \cellno 
& \cellno
& \cellno 
\\
\hline

& \sysName{$\mathsf{\sf \bf BeMap}$}~\cite{lin2023bemap}
&
& \cellyes 
& \cellno
& \cellno
& \cellno 
& \cellno
& \cellno 
& \cellno 
& \cellno
& \cellno
& \cellyes
& \cellno
& \cellno
& \cellyes
& \cellno 
& \cellno 
& \cellno
& \cellno
& \cellyes
& \cellno 
& \cellno
& \cellno 
\\
\hline

& \sysName{$\mathsf{\sf \bf CFC}$}~\cite{bose2019compositional}
&
& \cellyes 
& \cellyes
& \cellno
& \cellno 
& \cellno
& \cellno 
& \cellno 
& \cellno
& \cellyes
& \cellyes
& \cellno
& \cellno
& \cellno
& \cellyes 
& \cellno 
& \cellyes
& \cellno
& \cellno
& \cellno 
& \cellno
& \cellno 
\\
\hline

& \sysName{$\mathsf{\sf \bf FLIP}$}~\cite{masrour2020bursting}
&
& \cellyes
& \cellno
& \cellno
& \cellno 
& \cellno
& \cellno 
& \cellno 
& \cellno
& \cellno
& \cellno
& \cellno
& \cellno
& \cellno
& \cellyes
& \cellno 
& \cellyes
& \cellno
& \cellno
& \cellno 
& \cellno
& \cellno 
\\
\hline

& \sysName{$\mathsf{\sf \bf DKGE}$}~\cite{fisher2020debiasing}
&
& \cellno
& \cellno
& \cellno
& \cellyes 
& \cellno
& \cellno 
& \cellno 
& \cellno
& \cellno
& \cellno
& \cellno
& \cellno
& \cellno
& \cellyes 
& \cellno 
& \cellyes
& \cellno
& \cellno
& \cellno 
& \cellno
& \cellno 
\\
\hline

& \sysName{$\mathsf{\sf \bf FairGNN}$}~\cite{dai2021say}
&
& \cellyes
& \cellno
& \cellno
& \cellno
& \cellno
& \cellno 
& \cellno 
& \cellno
& \cellno
& \cellno
& \cellno
& \cellno
& \cellno
& \cellyes 
& \cellno 
& \cellyes
& \cellno
& \cellno
& \cellno 
& \cellno
& \cellno 
\\
\hline

& \sysName{$\mathsf{\sf \bf MONET}$}~\cite{palowitch2019monet}
&
& \cellyes
& \cellno
& \cellno
& \cellno
& \cellno
& \cellno 
& \cellno 
& \cellno
& \cellno
& \cellno
& \cellno
& \cellno
& \cellyes
& \cellno 
& \cellno 
& \cellno
& \cellno
& \cellyes
& \cellno 
& \cellno
& \cellno 
\\
\hline

& \sysName{$\mathsf{\sf \bf NIFTY}$}~\cite{agarwal2021towards}
&
& \cellyes
& \cellno
& \cellno
& \cellno
& \cellno
& \cellno 
& \cellno 
& \cellno
& \cellno
& \cellno
& \cellno
& \cellno
& \cellyes
& \cellno 
& \cellno 
& \cellno
& \cellno
& \cellyes
& \cellno 
& \cellno
& \cellno 
\\
\hline

& \sysName{$\mathsf{\sf \bf Crosswalk}$}~\cite{khajehnejad2021crosswalk}
&
& \cellyes
& \cellno
& \cellno
& \cellno 
& \cellno
& \cellno 
& \cellno 
& \cellyes
& \cellno
& \cellno
& \cellno
& \cellno
& \cellyes
& \cellno 
& \cellno 
& \cellyes
& \cellno
& \cellyes
& \cellno 
& \cellno
& \cellyes
\\
\hline

& \sysName{$\mathsf{\sf \bf InFoRM}$}~\cite{inform2020}
&
& \cellyes
& \cellno
& \cellno
& \cellno 
& \cellno
& \cellno 
& \cellno 
& \cellno
& \cellno
& \cellyes 
& \cellno
& \cellyes
& \cellyes
& \cellyes 
& \cellno 
& \cellno
& \cellno
& \cellno
& \cellno 
& \cellno
& \cellyes 
\\
\hline

& \sysName{$\mathsf{\sf \bf REDRESS}$}~\cite{dong2021individual}
&
& \cellyes
& \cellno
& \cellno
& \cellno 
& \cellno
& \cellno 
& \cellno 
& \cellno
& \cellno
& \cellyes 
& \cellno
& \cellno
& \cellyes
& \cellno 
& \cellno 
& \cellyes
& \cellno
& \cellyes
& \cellno 
& \cellno
& \cellno 
\\
\hline

& \sysName{$\mathsf{\sf \bf FairHELP}$}~\cite{cao2023fairhelp}
&
& \cellno
& \cellno
& \cellyes
& \cellno
& \cellno
& \cellno 
& \cellno 
& \cellno
& \cellno
& \cellno
& \cellno
& \cellno
& \cellyes
& \cellno 
& \cellno 
& \cellyes
& \cellno
& \cellno
& \cellno 
& \cellno
& \cellno 
\\
\hline

& \sysName{$\mathsf{\sf \bf FAN}$}~\cite{arduini2020adversarial}
&
& \cellno 
& \cellno
& \cellno
& \cellyes
& \cellno
& \cellno 
& \cellno 
& \cellno
& \cellno
& \cellno
& \cellno
& \cellno
& \cellyes 
& \cellno 
& \cellno 
& \cellno
& \cellno
& \cellno
& \cellno 
& \cellno
& \cellno 
\\
\hline

& \sysName{$\mathsf{\sf \bf HyperGCL}$}~\cite{weihypergcl}
&
& \cellno 
& \cellno
& \cellno
& \cellno
& \cellyes
& \cellno 
& \cellno 
& \cellno
& \cellno
& \cellno
& \cellno
& \cellno
& \cellyes 
& \cellno 
& \cellno 
& \cellno
& \cellno
& \cellno
& \cellno 
& \cellno
& \cellno 
\\
\hline

& \sysName{$\mathsf{\sf \bf FairEvolveGCN}$}~\cite{song2022individual}
&
& \cellno
& \cellno
& \cellno
& \cellno 
& \cellno
& \cellyes 
& \cellno 
& \cellno
& \cellno
& \cellyes
& \cellno
& \cellno
& \cellyes
& \cellno 
& \cellno 
& \cellyes
& \cellno
& \cellno
& \cellno 
& \cellno
& \cellno 
\\
\hline

\noalign{\hrule height 0.7pt}
\end{tabular}
\end{center}
}

\end{minipage}
\end{table}
}

%% file: table-datasets.tex
\definecolor{googleblue}{RGB}{66,133,244}
\definecolor{googlered}{RGB}{219,68,55}
\definecolor{googlegreen}{RGB}{15,157,88}
\definecolor{googlepurple}{RGB}{138,43,226}
\definecolor{lightred}{RGB}{255,198,196}
\definecolor{lightblue}{RGB}{197, 241, 255}
\definecolor{lightgreen}{RGB}{200, 247, 200}
\definecolor{lightpurple}{RGB}{230,230,250}
\definecolor{lightyellow}{RGB}{242, 232, 99}
\definecolor{lighterblue}{RGB}{197, 220, 255}
\definecolor{lighterred}{RGB}{253, 249, 205}
\definecolor{lightyellow}{RGB}{207, 161, 13}

\begin{table}
\centering
\caption{
Summary of datasets used for fairness evaluation of GNNs.
Note $|V|$ and $|E|$ are the number of nodes and edges, respectively.
Further, $|S|$ is the number of unique values of the sensitive attribute $S$.
Also, $|\mathcal{S}|$ is the number of sensitive attributes 
and $|\mX|$ is the number of input features (if any).
We categorize datasets according to their domain (recommendation, social networks, collaboration, web graphs, similarity, or citation networks) along with the task that each dataset was used.
}
\vspace{-2.5mm}
\label{table:datasets}
\renewcommand{\arraystretch}{1.10} 
\small
\footnotesize
\begin{tabularx}{1.0\linewidth}{ll c c c H H cH p{22mm} @{}H @{}c@{}}
\toprule
& \textsc{Dataset} & $|V|$ & $|E|$ & $S$ ($|S|$) & $|S|$ & $|\mathcal{S}|$ & $|\mX|$ & & \textsc{Description} 
& 
& \textsc{Task}
\\
\bottomrule
\TTT
\cellcolor{lightred}\textcolor{googlered}{\large\textsc{rec}} & \cellcolor{lightred}\textsf{ML-100K} & \cellcolor{lightred}2,625 & \cellcolor{lightred}100K & \cellcolor{lightred}gender (2), age (7), job (21) & \cellcolor{lightred}7 & \cellcolor{lightred}3 & \cellcolor{lightred}12 & \cellcolor{lightred} & \cellcolor{lightred} user-by-movie & DeBayes & \cellcolor{lightred}LP \\ 

\cellcolor{lightred} & \cellcolor{lightred}\textsf{ML-1M} & \cellcolor{lightred}10,040 & \cellcolor{lightred}1M & \cellcolor{lightred}gender (2), age (7), job (21) & \cellcolor{lightred}7 & \cellcolor{lightred}3 & \cellcolor{lightred}11 & \cellcolor{lightred}& \cellcolor{lightred}user-by-movie & \cellcolor{lightred}
FairGO~\cite{wu2021learning} 
& \cellcolor{lightred}LP \\ 
\cellcolor{lightred}& \cellcolor{lightred}\textsf{LastFM} & \cellcolor{lightred}49,900 & \cellcolor{lightred}518,647 & \cellcolor{lightred}gender (2), age (3) & \cellcolor{lightred}2 & \cellcolor{lightred}1 & \cellcolor{lightred}$-$ & \cellcolor{lightred}& \cellcolor{lightred}user-by-artist \cellcolor{lightred} & \cellcolor{lightred}
FairGO~\cite{wu2021learning}, 
FairRec~\cite{patro2020fairrec} 
& \cellcolor{lightred}LP \\ 
\cellcolor{lightred} & \cellcolor{lightred}\textsf{Amazon} & \cellcolor{lightred}334,863 & \cellcolor{lightred}925,872 & \cellcolor{lightred}product category (4) & \cellcolor{lightred}2 & \cellcolor{lightred}1 & \cellcolor{lightred}$-$ & \cellcolor{lightred}& \cellcolor{lightred}user-by-product & \cellcolor{lightred}Method & \cellcolor{lightred}LP \\ 
\cellcolor{lightred} & \cellcolor{lightred}\textsf{Yelp} & \cellcolor{lightred}12,683 & \cellcolor{lightred}211,721 & \cellcolor{lightred}food genre (4) & \cellcolor{lightred}2 & \cellcolor{lightred}1 & \cellcolor{lightred}14 & \cellcolor{lightred}& \cellcolor{lightred}user-by-business & \cellcolor{lightred}Method & \cellcolor{lightred}LP \\ 
\hline

\cellcolor{lightblue}\textcolor{googleblue}{\large\textsc{social}} 
\cellcolor{lightblue}& \cellcolor{lightblue}\textsf{Pokec} & \cellcolor{lightblue}1.63M & \cellcolor{lightblue}30.6M & \cellcolor{lightblue}gender (2), region (2) & \cellcolor{lightblue}2 & \cellcolor{lightblue}1 & \cellcolor{lightblue}59 & \cellcolor{lightblue}& \cellcolor{lightblue}friendship & \cellcolor{lightblue}Method & \cellcolor{lightblue}LP \\ 
\cellcolor{lightblue}& \cellcolor{lightblue}\textsf{Pokec-n} & \cellcolor{lightblue}66,569 & \cellcolor{lightblue}729,129 & \cellcolor{lightblue}region (2) & \cellcolor{lightblue}2 & \cellcolor{lightblue}1 & \cellcolor{lightblue}59 & & \cellcolor{lightblue}friendship & \cellcolor{lightblue}Method & \cellcolor{lightblue}NC \\ 
\cellcolor{lightblue}& \cellcolor{lightblue}\textsf{Pokec-z} & \cellcolor{lightblue}67,797 & \cellcolor{lightblue}882,765 & \cellcolor{lightblue}region (2) & \cellcolor{lightblue}2 & \cellcolor{lightblue}1 & \cellcolor{lightblue}59 & \cellcolor{lightblue}& \cellcolor{lightblue}friendship & \cellcolor{lightblue}Method & \cellcolor{lightblue}NC \\ 
\cellcolor{lightblue}& \cellcolor{lightblue}\textsf{Twitter}$^{*}$ & \cellcolor{lightblue}81,306 & \cellcolor{lightblue}1,768,149 &\cellcolor{lightblue} political view (2) & \cellcolor{lightblue}2 & 1 & \cellcolor{lightblue}1,364 & \cellcolor{lightblue}& \cellcolor{lightblue}who-follows-whom & \cellcolor{lightblue}Method & \cellcolor{lightblue}LP \\ 
\cellcolor{lightblue}& \cellcolor{lightblue}\textsf{Facebook} & \cellcolor{lightblue}1,034 & \cellcolor{lightblue}26,749 & \cellcolor{lightblue}gender (2) & \cellcolor{lightblue}2 & \cellcolor{lightblue}1 & \cellcolor{lightblue}224 & & \cellcolor{lightblue}friendship & \cellcolor{lightblue}Method & \cellcolor{lightblue}LP \\ 
\cellcolor{lightblue}& \cellcolor{lightblue}\textsf{fb-Ok97} & \cellcolor{lightblue}3,111 & \cellcolor{lightblue}73,230 & \cellcolor{lightblue}gender (2) & \cellcolor{lightblue}2 & \cellcolor{lightblue}1 & \cellcolor{lightblue}8 & \cellcolor{lightblue}& \cellcolor{lightblue}friendship & \cellcolor{lightblue}Method & \cellcolor{lightblue}LP,NC \\ 
\cellcolor{lightblue}& \cellcolor{lightblue}\textsf{fb-UNC28} & \cellcolor{lightblue}4,018 & \cellcolor{lightblue}65,287 & \cellcolor{lightblue}gender (2) & \cellcolor{lightblue}2 & \cellcolor{lightblue}1 & \cellcolor{lightblue}8 & \cellcolor{lightblue}& \cellcolor{lightblue}friendship & \cellcolor{lightblue}Method & \cellcolor{lightblue}LP,NC \\ 
\cellcolor{lightblue}& \cellcolor{lightblue}\textsf{fb-Rice} & \cellcolor{lightblue}1,205 & \cellcolor{lightblue}42,443 & \cellcolor{lightblue}age (2) & \cellcolor{lightblue}2 & \cellcolor{lightblue}1 & \cellcolor{lightblue}2 & \cellcolor{lightblue}& \cellcolor{lightblue}friendship & CrossWalk & \cellcolor{lightblue}LP, NC \\
\cellcolor{lightblue}& \cellcolor{lightblue}\textsf{Google+} & \cellcolor{lightblue}4,938 & \cellcolor{lightblue}547,923 & \cellcolor{lightblue}gender (2) & \cellcolor{lightblue}2 & \cellcolor{lightblue}1 & \cellcolor{lightblue}5 & \cellcolor{lightblue}& \cellcolor{lightblue}friendship & Method & \cellcolor{lightblue}LP \\ 
\cellcolor{lightblue}& \cellcolor{lightblue}\textsf{NBA} & \cellcolor{lightblue}403 & \cellcolor{lightblue}10,621 & \cellcolor{lightblue}country (2) & \cellcolor{lightblue}2 & \cellcolor{lightblue}1 & \cellcolor{lightblue}96 & \cellcolor{lightblue}& \cellcolor{lightblue}who-follows-whom & NIFTY & \cellcolor{lightblue}NC \\ 
\cellcolor{lightblue}& \cellcolor{lightblue}\textsf{fb-Gender} & \cellcolor{lightblue}7,315 & \cellcolor{lightblue}89,733 & \cellcolor{lightblue}gender (2) & \cellcolor{lightblue}2 & 1 & \cellcolor{lightblue}$-$ & \cellcolor{lightblue}& \cellcolor{lightblue}friendship & \cellcolor{lightblue}FairNeigh & \cellcolor{lightblue}LP \\
\cellcolor{lightblue}& \cellcolor{lightblue}\textsf{Retweet-pol} & \cellcolor{lightblue}18,470 & \cellcolor{lightblue}61,157 & \cellcolor{lightblue}political view (2) & \cellcolor{lightblue}2 & \cellcolor{lightblue}1 & \cellcolor{lightblue}$-$ & \cellcolor{lightblue}& \cellcolor{lightblue}friendship & \cellcolor{lightblue}FairNeigh & \cellcolor{lightblue}LP \\
\cellcolor{lightblue}& \cellcolor{lightblue}\textsf{Dutch school} & \cellcolor{lightblue}26 & \cellcolor{lightblue}221 & \cellcolor{lightblue}gender (2) & 2 & \cellcolor{lightblue}1 & \cellcolor{lightblue}$-$ & & \cellcolor{lightblue}friendship & \cellcolor{lightblue}Method~\cite{masrour2020bursting} & \cellcolor{lightblue}LP \\ 
\cellcolor{lightblue}& \cellcolor{lightblue}\textsf{Epinion} & \cellcolor{lightblue}8,806 & \cellcolor{lightblue}157,887 & \cellcolor{lightblue}$-$ & \cellcolor{lightblue}2 & \cellcolor{lightblue}1 & \cellcolor{lightblue}$-$ & \cellcolor{lightblue}& \cellcolor{lightblue}user-trusts-user & \cellcolor{lightblue}Method & \cellcolor{lightblue}
LP
\\ 
\cellcolor{lightblue}& \cellcolor{lightblue}\textsf{Ciao} & \cellcolor{lightblue}7,317 & \cellcolor{lightblue}85,205 & \cellcolor{lightblue}$-$ & \cellcolor{lightblue}2 & \cellcolor{lightblue}1 & \cellcolor{lightblue}$-$ & \cellcolor{lightblue}& \cellcolor{lightblue}user-trusts-user & \cellcolor{lightblue}Method & \cellcolor{lightblue}LP
\\ 
\cellcolor{lightblue}& \cellcolor{lightblue}\textsf{Filmtrust} & \cellcolor{lightblue}3,579 & \cellcolor{lightblue}35,494 & \cellcolor{lightblue}$-$ & \cellcolor{lightblue}2 & \cellcolor{lightblue}1 & \cellcolor{lightblue}$-$ & \cellcolor{lightblue}& \cellcolor{lightblue}user-trusts-user & \cellcolor{lightblue}Method & \cellcolor{lightblue}
LP
\\ 
\hline
\multirow{1}{*}{\textcolor{googlegreen}{\large\textsc{collab}}}
\cellcolor{lightgreen}& \cellcolor{lightgreen} \cellcolor{lightgreen}\textsf{Citeseer} & \cellcolor{lightgreen} \cellcolor{lightgreen}3,327 & \cellcolor{lightgreen} 4,732 & \cellcolor{lightgreen} \cellcolor{lightgreen}topic (6) & \cellcolor{lightgreen}  & \cellcolor{lightgreen} 1 & \cellcolor{lightgreen} 3,703 & \cellcolor{lightgreen} & \cellcolor{lightgreen} coauthorship & \cellcolor{lightgreen} Method & \cellcolor{lightgreen} LP \\ 
\cellcolor{lightgreen}& \cellcolor{lightgreen} \textsf{Cora} & \cellcolor{lightgreen} 2,708 & \cellcolor{lightgreen} 5,429 & \cellcolor{lightgreen} topic (7) & \cellcolor{lightgreen} \cellcolor{lightgreen}5 & \cellcolor{lightgreen} \cellcolor{lightgreen}1 & \cellcolor{lightgreen} 1,433 & \cellcolor{lightgreen} & \cellcolor{lightgreen} coauthorship & \cellcolor{lightgreen} Method & \cellcolor{lightgreen} LP \\ 
\cellcolor{lightgreen}& \cellcolor{lightgreen} \textsf{Pubmed} & \cellcolor{lightgreen} 19,717 & \cellcolor{lightgreen} 44,338 & \cellcolor{lightgreen} topic (3) & \cellcolor{lightgreen} 5 & \cellcolor{lightgreen} 1 & \cellcolor{lightgreen} 500 & \cellcolor{lightgreen} & \cellcolor{lightgreen} coauthorship & \cellcolor{lightgreen} Method & \cellcolor{lightgreen} LP \\ 
\cellcolor{lightgreen}& \cellcolor{lightgreen} \textsf{DBLP} & \cellcolor{lightgreen} 3,980 & \cellcolor{lightgreen} 6,965 & \cellcolor{lightgreen} continent (5), gender(2) & \cellcolor{lightgreen} 5 & \cellcolor{lightgreen} 1 & \cellcolor{lightgreen} $-$ & \cellcolor{lightgreen} & \cellcolor{lightgreen} coauthorship & \cellcolor{lightgreen} DeBayes & \cellcolor{lightgreen} LP \\ 
\cellcolor{lightgreen}& \cellcolor{lightgreen} \textsf{Hospital-cont}
\cellcolor{lightgreen}& \cellcolor{lightgreen} \cellcolor{lightgreen}75 & \cellcolor{lightgreen} \cellcolor{lightgreen}1139 & \cellcolor{lightgreen} \cellcolor{lightgreen}job (4) & \cellcolor{lightgreen} 2 & \cellcolor{lightgreen} 1 & \cellcolor{lightgreen} $-$ & \cellcolor{lightgreen} & \cellcolor{lightgreen} 
proximity & \cellcolor{lightgreen} 
Method~\cite{masrour2020bursting} & \cellcolor{lightgreen} LC \\
\hline
\multirow{1}{*}{\textcolor{googlepurple}{\large\textsc{web}}}
\cellcolor{lightpurple}& \cellcolor{lightpurple}  \textsf{WebKB} & \cellcolor{lightpurple} 265 & \cellcolor{lightpurple} 530 & \cellcolor{lightpurple} topic (5) & \cellcolor{lightpurple} 5 & \cellcolor{lightpurple} 1 & \cellcolor{lightpurple} 500 & \cellcolor{lightpurple} & \cellcolor{lightpurple} web graph & \cellcolor{lightpurple} FairNeigh & \cellcolor{lightpurple} LP \\ 
\cellcolor{lightpurple}& \cellcolor{lightpurple} \textsf{Chameleon} & \cellcolor{lightpurple} 2,277 & \cellcolor{lightpurple} 31,371 & \cellcolor{lightpurple} gen. node degree (2) & \cellcolor{lightpurple} 5 & \cellcolor{lightpurple} 1 & \cellcolor{lightpurple} 2,325 & \cellcolor{lightpurple} & \cellcolor{lightpurple} web graph & \cellcolor{lightpurple} DegFairGCN & \cellcolor{lightpurple} NC \\ 
\cellcolor{lightpurple}& \cellcolor{lightpurple} \textsf{Squirrel} & \cellcolor{lightpurple} 5,201 & \cellcolor{lightpurple} 198,353 & \cellcolor{lightpurple} gen. node degree (2) & \cellcolor{lightpurple} 5 & \cellcolor{lightpurple} 1 & \cellcolor{lightpurple} 2,089 & \cellcolor{lightpurple} & \cellcolor{lightpurple} web graph & \cellcolor{lightpurple} DegFairGCN & \cellcolor{lightpurple} NC \\ 
\hline
%
\textcolor{darkblue}{\large\textsc{sim}} 
\cellcolor{lighterblue}& \cellcolor{lighterblue} \textsf{German} & \cellcolor{lighterblue} 1,000 & \cellcolor{lighterblue} 21,742 & \cellcolor{lighterblue} gender (2) & \cellcolor{lighterblue} 2 & \cellcolor{lighterblue} 1 & \cellcolor{lighterblue} 27 & \cellcolor{lighterblue} & \cellcolor{lighterblue} client similarity & \cellcolor{lighterblue} NIFTY & \cellcolor{lighterblue} NC \\ 
\cellcolor{lighterblue}& \cellcolor{lighterblue} \textsf{Recidivism} & \cellcolor{lighterblue} 18,876 & \cellcolor{lighterblue} 311,870 & \cellcolor{lighterblue} race (2) & \cellcolor{lighterblue} 2 & \cellcolor{lighterblue} 1 & \cellcolor{lighterblue} 18 & \cellcolor{lighterblue} & \cellcolor{lighterblue} defendant sim. & \cellcolor{lighterblue} Method & \cellcolor{lighterblue} NC \\ 
\cellcolor{lighterblue}& \cellcolor{lighterblue} \textsf{Credit} & \cellcolor{lighterblue} 30,000 & \cellcolor{lighterblue} 1,421,858 & \cellcolor{lighterblue} age (2) & \cellcolor{lighterblue} 2 & \cellcolor{lighterblue} 1 & \cellcolor{lighterblue} 13 & \cellcolor{lighterblue} & \cellcolor{lighterblue} individual sim. & \cellcolor{lighterblue} NIFTY & \cellcolor{lighterblue} NC \\ 
\hline
\textcolor{lightyellow}{\large\textsc{cit}} 
\cellcolor{lighterred}& \cellcolor{lighterred}\textsf{EMNLP} & \cellcolor{lighterred}2,600 & \cellcolor{lighterred}7,969 & \cellcolor{lighterred}gen. node degree (2) &\cellcolor{lighterred} 5 &\cellcolor{lighterred} 1 & \cellcolor{lighterred}8 &\cellcolor{lighterred} & \cellcolor{lighterred}citation network & \cellcolor{lighterred}DegFairGCN & \cellcolor{lighterred}\cellcolor{lighterred}NC \\ 
\hline
\end{tabularx}
\end{table}